\documentclass{article}
\usepackage{log_2024}

\usepackage{booktabs}						%
\usepackage{multirow}						%
\usepackage{amsfonts}						%
\usepackage{graphicx}						%
\usepackage{duckuments}						%
\usepackage{caption}
\usepackage{booktabs} %
\usepackage{amsmath}
\usepackage{microtype}
\usepackage{amssymb,graphicx,color}
\usepackage{amsfonts}
\usepackage{latexsym}
\usepackage{subcaption}
\usepackage{amssymb}
\usepackage{mathtools}
\usepackage{algorithm}
\usepackage{algorithmic}
\usepackage{wrapfig}

\newcommand\showchanges{0}

\newcommand{\chl}[1]{%
    \ifnum 1=\showchanges \relax
        {\color{orange}#1}\else #1%
    \fi
}

\usepackage[numbers,compress,sort]{natbib}	%

\title[Reinforcement Learning Discovers Efficient Decentralized Graph Path Search Strategies]{Reinforcement Learning Discovers Efficient Decentralized Graph Path Search Strategies}

\author[A. Pisacane et al.]{%
Alexei Pisacane\thanks{Correspondence to \texttt{pisacane.alexei@gmail.com}.}\hspace{0.125em} \textsuperscript{1}, Victor-Alexandru Darvariu\textsuperscript{2}, Mirco Musolesi\textsuperscript{1,3}\\
{\textsuperscript{1}University College London \quad \textsuperscript{2}University of Oxford \quad \textsuperscript{3}University of Bologna}\\
}

\begin{document}

\maketitle

\begin{abstract}
Graph path search is a classic computer science problem that has been recently approached with Reinforcement Learning (RL) due to its potential to outperform prior methods. Existing RL techniques typically assume a global view of the network, which is not suitable for large-scale, dynamic, and privacy-sensitive settings. An area of particular interest is search in social networks due to its numerous applications. 
Inspired by seminal work in experimental sociology, which showed that decentralized yet efficient search is possible in social networks, we frame the problem as a collaborative task between multiple agents equipped with a limited local view of the network. We propose a multi-agent approach for graph path search that successfully leverages both homophily and structural heterogeneity. Our experiments, carried out over synthetic and real-world social networks, demonstrate that our model significantly outperforms learned and heuristic baselines. Furthermore, our results show that meaningful embeddings for graph navigation can be constructed using reward-driven learning.
\end{abstract}

\section{Introduction}
\label{Introduction}
Graph path search is a fundamental task in Computer Science, pivotal in various domains such as knowledge bases~\cite{suchanek2007yago}, robotics~\cite{lavalle2006planning}, and social networks~\cite{doi:10.1126/science.1070120}. Given a start node and end node, the goal is to find a path from a source to a destination in the graph that connects them and optimizes desiderata such as minimizing path length. We refer to search strategies that achieve this as \textit{efficient}.
The problem is generally framed from a centralized perspective with a global view of the network, which is impractical or infeasible for several applications. In peer-to-peer networks~\cite{oram2001peer}, where privacy is a primary concern, a centralized agent poses significant risks~\cite{albert2000error,aspnes2002fault,callaway2000network}. Large graphs may also induce scalability bottlenecks as the storage requirements of a centralized directory strain memory limitations~\cite{chawathe2003making}. Moreover, in dynamic networks, maintaining a consistent global view of the topology may be impossible~\cite{giordano2002mobile}. 
Graph path search is of particular interest in social networks given the inherent commercial applications and potential for new insights from a social sciences perspective \cite{easley2010networks,wasserman1994social}.

In this paper, we will study the problem of decentralized path graph search using local information.
We will consider social networks and we will discuss how the proposed method can be directly applied to any networks for which topological and node attribute information is available. 
Indeed, prior experiments in human social networks, such as Stanley Milgram's renowned ``small world" experiment \cite{travers1969}\footnote{Milgram's experiment investigated the degree of connectedness among people in the United States, leading to the concept of ``six degrees of separation'',
the idea that any two people on Earth are connected by a chain of no more than six acquaintances, which has also entered popular culture \cite{guare2016six}. Milgram selected participants from Nebraska and Kansas. Each participant was given a letter and instructed to send it to a target person, a stockbroker living in Boston. However, they could only send the letter to someone they knew personally, characterized by certain (node) \textit{attributes} who they thought might be closer to the target. Each recipient of the letter would then forward it to someone they knew personally, continuing this process until the letter reached the target.
On average, it took about six steps for the letters to reach the stockbroker in Boston \cite{watts2004six}.}  reveals the existence of short paths in social networks that are discoverable solely through local graph topology and high-level node attributes, e.g., characteristics of the individuals, such as their occupation, their high-school, and so on. Many social networks exhibit two key properties that make decentralized search with partial information possible and efficient. The first, \emph{homophily} \cite{mcpherson2001birds}, reflects the tendency of individuals to connect with others who share similar attributes. The second is the \textit{heterogeneity} of local structure in many networks, in which nodes are often organized into highly connected communities \cite{fortunato2010community}, with a smaller number of \emph{weak ties} \cite{granovetter1973strength,aral2016future} or central connectors \cite{newman2003structure} bridging these node clusters and acting as shortcuts. 

Reinforcement Learning (RL) has recently been employed with a centralized perspective for discovering learned heuristics for graph search~\cite{bhardwaj2017learning,pandy2022learning} and reasoning over paths in knowledge graphs~\cite{das2018go,shen2018m}, in a way that complements or outperforms classic algorithms. 
Motivated by the promise of RL and the goal to attain decentralized graph path search, in this paper, we propose a multi-agent RL formulation in the Decentralized Partially Observable Markov Decision Process (Dec-POMDP) framework. 
Agents have \textit{only local visibility of graph topology and neighbor attributes}, and cooperate towards finding paths to the target node. We propose a method for learning in this Dec-POMDP that, in accordance with the Centralized Training and Decentralized Execution (CTDE) paradigm~\cite{Foerster_Farquhar_Afouras_Nardelli_Whiteson_2018}, trains an actor-critic model with node representations learned by a Graph Attention Network~\cite{velivckovic2017graph} with shared parameters. 
These embeddings are computed via a message-passing procedure starting from the raw node attributes and the graph topology. For this reason we name the resulting method \textbf{GARDEN}: \textbf{G}raph \textbf{A}ttention-guided \textbf{R}outing for \textbf{DE}centralised \textbf{N}etworks.
At execution time, the policy is used in a decentralized fashion by all agents.

We conduct experiments on synthetic and real-world social network graphs with up to 600 nodes to evaluate our model design. Our findings highlight the superior ability of our method to utilize both homophily and local connectivity better when compared to learned and handcrafted baseline models. Moreover, we find that the learned embeddings are meaningful representations for navigation in high-dimensional feature space.
Our results show that the dynamics observed in Milgram's experiment can emerge using reward-driven learning. RL is able to construct a latent feature space that effectively incorporates both node attribute information and network structure for use in graph path search. Therefore, this work supports the notion that decentralized graph path search can succeed given appropriate representations, and shows a possible mechanism for how representations similar to those inherently used by individuals may be constructed.

\section{Related Work}\label{relatedwork}
\subsection{Network Science}\label{subsec:netsci}

Search is a common operation in network applications. Various classic algorithms~\cite{cormen2022introduction} ensure path discovery between two nodes under specific conditions. They require maintaining global knowledge of the graph structure, which, as we have argued, is impractical in certain cases due to considerations of privacy, scalability, and dynamicity. We therefore focus our attention on graph path search using only local information.

As previously discussed, our inspiration for studying this problem is Milgram's ``small world" experiment~\cite{travers1969}. The findings, later validated on a larger scale~\cite{dodds2003experimental}, support this hypothesis in social networks, which are characterized by short mean path lengths. Subsequent research~\cite{killworth1978reversal} highlighted the discovery of effective routing strategies, emphasising the concept of homophily \cite{mcpherson2001birds}, which states that individuals seek connections to others that are similar to themselves.

In addition to homophily, many networks are characterized by a power-law degree distribution~\cite{barabasi1999emergence} and exhibit heterogeneity in node degree. In such networks, a few ``hub" nodes with numerous connections coexist with many nodes having a relatively small degree. Highly connected nodes therefore offer potential shortcuts in search trajectories by bridging sparsely connected communities.

For effective search, finding the bridging node between two communities is often required. Relying solely on homophily or node degree may be ineffective, as the bridging node might lack a large degree or significant attribute similarity with the target node. In networks with large clusters, an agent may spend considerable time navigating the current cluster before reaching the desired community. Identifying \textit{weak ties} \cite{granovetter1973strength} between communities is challenging using only node attributes or degrees; therefore, an effective search for weak ties requires awareness of candidate nodes' neighborhoods.

A useful lens for viewing this problem is through a ``hidden metric space"~\cite{boguna2009navigability} of node features. Assuming node features are representative of their position within this space, the probability of edge-sharing increases with decreasing pairwise attribute distance. Empirical evidence supports the efficiency of navigation using this underlying metric. If an approximation to the hidden metric using only local graph structure is feasible, a decentralized strategy could involve moving toward nodes minimizing the approximate metric~\cite{boguna2009navigability, kleinberg2000small}.

The approach of~\citet{simsek2005decentralized} is most closely related to ours as it also treats the problem of search by leveraging both homophily and the node degree disparity. The algorithm uses an estimate of the statistical relationship between the attribute similarity and connection probability whose computation requires knowledge of the attributes of all the nodes in the network. In contrast, our method does not require the availability of this global information. 

\subsection{Reinforcement Learning for Graph Routing and Search}

Reinforcement Learning methods have been applied for a variety of graph optimization problems in recent years as a mechanism for discovering, by trial-and-error, algorithms that can outperform classic heuristics and metaheuristics~\cite{darvariu2024graph}. Their appeal stems from the flexibility of the RL framework to address a wide variety of problems, requiring merely that they can be formulated as MDPs. The most relevant works in this area treat routing and search problems over graphs.

\chl{Early work on RL for routing demonstrated the potential of the MDP formalism~\citep{boyan1993packet,choi1995predictive, peshkin2002reinforcement}, but suffered from the main pitfall of tabular RL methods: poor scalability. Interest has been reignited recently by several works that employ function approximation for scaling to larger problems.} In this line of work,~\citet{valadarsky2017learning} considered learning routing policies on a dataset of historical traces of pairwise demands and applying them in new traffic conditions. The MDP is framed as learning a set of edge weights from which the routing strategy is determined.\chl{~\citet{hope2021gddr} expanded on this work by introducing a Graph Neural Network (GNN)~\cite{hamilton2020graph} technique for function approximation, showing its advantages over using simple feedforward neural networks.} More recent work by~\citet{ALMASAN2022184} leveraged a GNN representation trained using a policy gradient algorithm. They frame actions as the choice of a middle-point for a flow given start and target nodes, with previous action choices becoming part of the state. \chl{Other recent works on RL for routing considered optimizing a weighted combination of delay and throughput~\citep{xu2018experience} and deciding how to re-route the most important flows (i.e., those with the most traffic) given an initial routing scheme~\citep{zhang2020cfr}.}

Another important line of work studies how to perform search on graphs. In contrast to routing, for search tasks there is no notion of a link load associated with traversing a particular node or edge in the graph. A notable contribution in this direction is work by~\citet{pandy2022learning}, where RL agents are tasked with learning a heuristic function for augmentation of A* search.\chl{~\citet{patankar2023intrinsically} considered the task of validating the way in which humans perform graph navigation, adopting two theories relating topological graph properties to minimizing gaps in knowledge or compressing existing knowledge. Their DQN-based agent parameterized by a GNN was validated successfully using human graph navigation trajectories.}

The problem of \textit{knowledge graph completion} may also be viewed as graph traversal in instances with heterogeneous edge types~\cite{guu2015traversing} and with a target node that is not specified a priori.~\citet{das2018go} proposed an MDP formulation of this task, in which an agent chooses the next relationship to traverse given the current node. A proportion of the true relationships in the knowledge graph is masked and used to provide the reward signal for training the agent via REINFORCE.
The M-Walk method~\cite{shen2018m} builds further in this direction by leveraging the determinism of the transition dynamics. Therefore, training with trajectories from a Monte Carlo Tree Search (MCTS) policy~\cite{browne2012survey} can overcome the reward sparsity associated with the random exploration of model-free methods.\chl{~\citet{zhang2022learning} proposed a hierarchical method that features a high-level agent for choosing a cluster in which the target may be located, and a lower-level agent that navigates within the cluster.}

Lastly, we note that, while the works reviewed in this section share features of our MDP and model design, none are directly applicable to the problem formulation. Chiefly, we consider a decentralized graph path search scenario in which each agent has only partial visibility of the network.

\section{Methods}
In this section, we first introduce our decentralized mathematical formulation of the graph search problem. Next, we describe the proposed multi-agent reinforcement learning algorithm, which leverages learnable graph embeddings.

\subsection{MDP Formulation}

We frame the search problem as a Decentralized Partially Observable Markov Decision Process (Dec-POMDP)~\cite{POMDPbook} taking place over an attributed, undirected, and unweighted graph structure $G = (V,E)$ with $n$ nodes and $m$ edges. An agent is placed on each node $u_i \in V$ in the graph, while the edges $E$ indicate direct bidirectional communication links between agents. An attribute vector $\mathbf{x}_{u_i} \in \mathbb{R}^d$ is associated with each agent. The aim is to find a path starting from an initial node $u_{\text{src}}$ to a designated target node $u_{\text{tgt}}$ by passing a single global \textit{message}. 

\chl{
\textbf{States.} The global state $S_t$ at time $t$ is a tuple $\langle S_t^{(1)}, S_t^{(2)}, \dots, S_t^{(n)} \rangle$ composed of the states $S_{t}^{(i)} = (M_{t}^{(i)}, u_{\text{tgt}})$ of the individual agents. Here, $M_{t}^{(i)}$ is an indicator variable that denotes the presence or absence of the \textit{message} at a given node $u_{i}$ at time $t$. While specifying the target node $u_{\text{tgt}}$ is required for MDP stationarity, its identity is not provided to the agent in the observation.

\textbf{Actions.} The joint action space $\mathcal{A} = \bigtimes_{i} \mathcal{A}^{(i)}$ is the product of the agent-wise action spaces. At each timestep $t$ of an MDP episode, a node $u_{i}$ receives the \textit{message} $\mathbf{m} \in \mathbb{R}^d$ specifying the attributes $\mathbf{x}_{u_{\text{tgt}}}$ of the target node (but not its identity). It chooses as its action $A_t^{(i)}$ one of its neighbors, denoted $\mathcal{N}(u_{i})$, to pass the message on to. \chl{ We denote this action of node $u_{i}$ passing the message to node $u_{j}$ by $a_{u_{i}\to u_{j}}$.} All other agents take a no-op action at this step, which has no effect. Hence, $\mathcal{A}_{t}^{(i)} = \{a_{u_{i}\to u_{j}} | u_{j} \in \mathcal{N}(u_{i})\}$ if $M_{t}^{(i)} = 1$, and $\{\texttt{no-op}\}$ otherwise.

\textbf{Observations.} The environment emits a global observation $O_t = \langle O_t^{(1)}, O_t^{(2)}, \dots, O_t^{(n)} \rangle$ at each time step $t$, from which each node $u_i$ only observes its own component $O_{t}^{(i)}$.  In accordance with our motivations, we provide agents with only local observations of the graph topology. Concretely, we equip each agent $u_{i}$ with observations of 1-hop ego subgraphs $G_{u_{j}}$ centered on its neighboring nodes, including visibility of pairwise edges between 1-hop neighbors. The observation will also contain information on the target node: if the agent possesses the message, it symmetrically can observe an ego graph centered on $u_{\text{tgt}}$. Formally, the observation $O_{t}^{(i)}$ is defined as $(\mathbf{m}, \{G_{u_j} | u_j \in \mathcal{N}(u_{i}), G_{u_{\text{tgt}}}\})$ if $M_{t}^{(i)} = 1$, and $\{ G_{u_j} | u_j \in \mathcal{N}(u_{i}) \}$ otherwise. The ego graph $G_{u_j} = (V_{u_j}, E_{u_j})$ is defined such that $V_{u_j} = \mathcal{N}(u_j) \cup \{u_j\}$ and  $E_{u_j} = \{(u_k, u_l) \in E | u_k \in V_{u_j} \land u_l \in V_{u_j} \}$.

\textbf{Transitions.}  The message moves deterministically to the selected node, updating the indicator $M_{t+1}^{(i)}$ accordingly. Concretely, $M_{t+1}^{(i)} = 1$ if $M_{t}^{(j)} = 1 \land A_{t}^{(j)} = a_{u_{j} \to u_{i}}$, and $0$ otherwise. 

\textbf{Rewards}. The episode ends when the message reaches the target node, yielding a collective reward of +1 for the agents and terminating the episode. Formally, $R_{t+1}^{(k)} = 1\ \forall k$ if $\exists i\ .\ M_{t}^{(i)}=1 \land A_{t}^{(i)} = a_{u_{i} \to u_{\text{tgt}}}$, and $0$ otherwise. To prevent agents from entering action cycles, we introduce a truncation criterion: the episode can also end after $T_\text{max}$ interaction steps with the environment. 

}

\subsection{\chl{Learning Architecture}}\label{subsec:model}

In our design, we employ the common multi-agent Reinforcement Learning paradigm of Centralized Training with Decentralized Execution (CTDE) \cite{lowe2017multi}. We consider a fully collaborative setting in which the agents are all rewarded if messages are successfully delivered to the target node. The collaborative objective is formulated such that each agent selecting the optimal next action results in an optimal trajectory through the graph. Therefore, the optimal trajectory can be constructed in a decentralized manner. 

As it is common in the CTDE paradigm, we utilize parameter-sharing across agent networks. In the training scheme, a centralized agent receives localized observations from individual agents at each step, and is tasked with selecting optimal actions in the search path. The optimal decision is first learnt, and then replicated and distributed to individual nodes at execution time. 

At each training step, a central agent is given incomplete observations and receives sparse and delayed rewards from the environment. Given these specifications, we propose the use of a variant of the Advantage Actor-Critic (A2C) algorithm \cite{mnih2016asynchronous} to promote adequate exploration with acceptable sample efficiency. The A2C value network is also learned in a centralized fashion to guide the training of the policy network. Learning a stochastic policy (rather than a deterministic one) is important for the problem under consideration given that a short path to the target may not be available via a particular neighbor despite a high level of attribute similarity. Lastly, we incorporate entropy regularization to ensure the policy maintains a high degree of randomness while still aiming to maximize the expected discounted return.

\chl{The goal is to learn, for each agent, a policy $\pi_{u_{i}}$ that maps observations to actions.} We formulate the choice of neighbor to which the message should be transmitted based on values output by an MLP-parameterized policy network \chl{$f_{\pi}^{\Theta_{1}}$}. The policy network is applied for each neighbor $u_j \in \mathcal{N}(u_i)$ of the node $u_i$ that is currently in possession of the message at time $t$, and the SoftMax function is used to derive a probability distribution. Concretely, for node $u_{i}$ in posession of the message $\mathbf{m}$, the policy is defined as:

\begin{align}
    \chl{\pi_{u_{i}}(a_{u_{i}\to u_{j}} | O^{(i)}) = \frac{\text{exp}(f_{\pi}^{\Theta_{1}}([\mathbf{x}_{u_j} || \mathbf{x}_{u_{\text{tgt}}}]))} {\sum_{u_k \in \mathcal{N}(u_i)}{\text{exp} (f_{\pi}^\chl{{\Theta_{1}}}([\mathbf{x}_{u_k} || \mathbf{x}_{u_{\text{tgt}}}]))}},}
\end{align}

where $[\cdot || \cdot]$ denotes concatenation. Similarly, to estimate the value function, we pass the current node $u_i$ and the target node $u_{\text{tgt}}$ attributes through an MLP-parameterized value network \chl{$f_{v}^{\Theta_{2}}$}:

\chl{\begin{align}
v(O^{(i)}) = f_{v}^{\Theta_{2}}([\mathbf{x}_{u_i} || \mathbf{x}_{u_{\text{tgt}}}]),
\end{align}}

Where $u_{i}$ is the node in possession of the message $\mathbf{m}$ at time $t$.
It is interesting to note that the node features that are used as input to the policy and value networks will impact the effectiveness of the learned policies. The simplest choice is to use the raw node features $\mathbf{x}_{u_i}$, and we denote the resulting algorithm as \textsc{MLPA2C}. We also consider the simplest extension to this model that minimally incorporates local graph topology by augmenting node attributes with node degrees, i.e., $\mathbf{x}_{u_i}^{\text{WD}} = [\mathbf{x}_{u_i} || \text{deg}(u_i)]$. We refer to this as \textsc{MLPA2CWD}.

\subsection{GARDEN}

Recall the ``hidden metric" hypothesis discussed in Section~\ref{subsec:netsci}, which posits that a viable policy can be motivated by moving through the graph to reduce node distance, provided a good approximation of the underlying metric is obtained. Instead of prescribing that the raw node attributes should be used to approximate this metric, we propose that relevant node features, which capture the potentially complex interplay between attributes and topologies, \textit{can be learned}. 
To do so, we suggest replacing raw node attributes with learned embeddings $\mathbf{x}_{u_i}^{\text{GAT}}$ obtained from a Graph Attention Network (GAT) \cite{velivckovic2017graph}, denoted \chl{$f_{\text{rep}}^{\Theta_{3}}$}. These embeddings are computed via a message-passing procedure starting from the raw node attributes and the graph topology. 

The method, which we refer to as Graph Attention-guided Routing for Decentralized Networks (GARDEN),
is shown using pseudocode in Algorithm~\ref{alg:MLPA2C}. \chl{The full set of model parameters $\Theta = \{ \Theta_{i} \}_{i=1}^{3}$
is trained implicitly as we take gradient descent steps over the combined episodic loss $\sum_{t} L_{t}^{(\pi)} + L_{t}^{(v)}$}. The node embeddings are recalculated at the start of each episode. The notation $[[\cdot]]$ denotes the partial stopping of gradients, and $H(p)$ denotes the entropy of a discrete distribution, given by $\sum_{i} p_{i} \text{log}(p_{i})$.

\begin{algorithm}[tb]
   \caption{Graph Attention-guided Routing for DEcentralized Networks (GARDEN).}
   \label{alg:MLPA2C}
\begin{algorithmic}[1]
   \STATE {\bfseries Input:} 
   Policy Network \chl{$f_{\pi}^{\Theta_{1}}$}, Value Network \chl{$f_{v}^{\Theta_{2}}$}, Graph Representation Network \chl{$f_{\text{rep}}^{\Theta_{3}}$}, Ego Graphs $\{G_{u} | u \in V \}$, entropy regularization coefficient $\lambda$, discount factor $\gamma$.
   \STATE {\bfseries Output:}
   Learned policy, value and representation networks $\chl{f_{\pi}^{\Theta_{1}}, f_{v}^{\Theta_{2}}, f_{\text{rep}}^{\Theta_{3}}}$.
   \chl{\STATE Randomly initialize model parameters $\Theta = \{ \Theta_{i} \}_{i=1}^{3}$}.
   \FOR{$i=1$ {\bfseries to} $N_{\text{episodes}}$}
       \STATE Initialize episode buffer $\mathcal{B}$
       \STATE Sample starting node $u$, target node $u_{\text{tgt}}$

       \FOR{$w \in V$}
            \STATE Compute $\mathbf{x}_w^{\text{GAT}}$ using \chl{$f_{\text{rep}}^{\Theta_{3}}$}
       \ENDFOR
       \STATE  $\mathbf{m} = \mathbf{x}_{u_{\text{tgt}}}^{\text{GAT}}$\;
       
       \STATE $t = 0$
       
       \WHILE{ $u \neq u_\text{tgt}$ and $t < T_\text{max} $}
            \STATE Sample action $a_{u\to u'} \sim \pi_{u}(\cdot \text{ } | \mathbf{m})$
            \STATE Move message to node $u'$, observe reward $r$
            \STATE Store transition $(u, u', r)$ in $\mathcal{B}$
            \STATE $u \leftarrow u'$ 
            \STATE $t = t+1$
       \ENDWHILE
       \STATE Initialize episode loss $L=0$
       \FOR{$(u, u', r)$ in $\mathcal{B}$}
            \STATE $\hat{A} = r + [[\chl{f_{v}^{\Theta_{2}}}([\mathbf{x}_{u'}^{\text{GAT}} || \mathbf{m}])]] - \chl{f_{v}^{\Theta_{2}}}([\mathbf{x}_{u}^{\text{GAT}} || \mathbf{m}])$
             \STATE $L^{(\pi)}=- \hat{A} \log{\chl{f_{\pi}^{\Theta_{1}}}([\mathbf{x}_{u'}^{\text{GAT}} || \mathbf{m}])} - \lambda H(\pi(u | \mathbf{m}))$\;
            \STATE $L^{(v)} = \hat{A}^2$
            \STATE $L = L + L^{(\pi)} + L^{(v)}$
       \ENDFOR
       \STATE Take gradient descent step on L w.r.t. \chl{$\Theta$ }
       \ENDFOR
\end{algorithmic}
\end{algorithm}

\section{Experimental Setup}
\subsection{Datasets}

\textbf{Real-world Graphs.} To assess the performance of decentralized graph strategies on real-world data, we consider several ego graphs from the Facebook social network~\cite{leskovec2012learning} present in the SNAP~\cite{snapnets} repository. These graphs depict individuals and their Facebook friendships. Each node is equipped with binary, anonymized attributes collected through surveys. Due to computational budget constraints, we select the largest connected components of 5 graphs such that they have between 100 and 600 nodes. High-level descriptive statistics for these graphs are presented in Table \ref{tab:FBMetrics} in the Appendix.

\textbf{Synthetic Graphs.} We additionally consider synthetically generated graphs that are both attributed and display homophily. This allows for the creation of a diverse range of graphs with varying degrees of sparsity, enabling evaluations under different synthetic conditions. We follow the generative graph construction procedure proposed by~\citet{kaiser2004spatial}, which samples node attributes uniformly from a unit box $[0,1]^{d}$ and creates edges stochastically according to the rule $p((u, u') \in E) = \max(1, \beta e^{-\alpha \|\mathbf{x}_{u} - \mathbf{x}_{u'}\|_{2}})$, where $\alpha$ and $\beta$ are scaling coefficients.

\subsection{Baselines}

The learned baselines we use are the MLPA2C and MLPA2CWD techniques as introduced in Section~\ref{subsec:model}. Furthermore, we consider a suite of heuristic baselines that utilize homophily and graph structure for graph path search. The simplest baseline, \textit{GreedyWalker}, selects the next node greedily based on the smallest Euclidean attribute distance: $\pi(u) = \text{argmin}_{u' \in \mathcal{N}(u)} \|\mathbf{x}_{u'} - \mathbf{x}_{u_{\text{tgt}}} \|_{2}.$

Given that deterministic policies may result in action loops, we generalize this to a stochastic agent (\textit{DistanceWalker}) that  acts via a SoftMax policy over attribute distances with a tuned temperature parameter: $\pi^{(\tau)}(u' | u) = \frac{\text{exp}(-\|\mathbf{x}_{u'} - \mathbf{x}_{u}\|_{2} / \tau)}{\sum_{u'' \in \mathcal{N}(u)} \text{exp}(-\|\mathbf{x}_{u''} - \mathbf{x}_{u}\|_{2} / \tau)}.$
Similarly, we consider the stochastic \textit{ConnectionWalker} agent, which uses a SoftMax policy over node degree: $\pi^{(\tau)}(u' | u) = \frac{\text{exp}(\text{deg}(u') / \tau)} {\sum_{u'' \in \mathcal{N}(u)} \text{exp}(\text{deg}(u'') / \tau)}.$
Lastly, the \textit{RandomWalker} baseline selects uniformly at random between nodes from the current neighbourhood: $\pi(u' | u) = \frac{1}{\text{deg}(u)}.$

The stochastic agents use a temperature parameter $\tau$ to control greediness. To perform a fair comparison with our learned models, we individually tune the temperature for the DistanceWalker and ConnectionWalker models using a validation set for each graph.

\subsection{Model Evaluation \& Selection}

For evaluating models, we consider the following metrics:
\begin{enumerate}
    \item Mean Oracle Ratio $\bar{R}_{\text{oracle}}$: the ratio between episode length and the shortest path length averaged over all source-destination pairs;
    \item Truncation Rate $R_\text{trunc}$: \% of episodes exceeding the truncation length $T_\text{max}$;
    \item Win Rate $R_\text{win}$: \% of episodes where a given agent obtains the relative shortest path length, with ties broken randomly.
\end{enumerate}   

To mitigate potential memorization of routes during training, especially when nodes are uniquely identifiable based on attributes, we partition the node set $V$ into three disjoint groups: $V^{\text{train}}$, $V^{\text{val}}$, and $V^{\text{test}}$ at ratios of $80\%/10\%/10\%$. The source node $u_\text{src}$ is sampled uniformly at random from $V$, while the target node $u_\text{tgt}$ depends on whether training, validation, or testing is performed. \chl{This ensures that the agent cannot memorize the path to a target node since, by construction, it is not encountered during training.} We always sample a ``fresh'' source-target pair during training, while for validation and evaluation the source-target pairs are serialized and stored (such that the performance evaluated over them is consistent). The Mean Oracle Ratio $\bar{R}_{\text{oracle}}$ is used as the primary metric for model validation and evaluation.

\subsection{Sensitivity Analysis of Graph Density Parameter}

Given a constant graph size, reduced graph density diminishes available paths to a target~\cite{lovasz1993random}. This intensifies exploration challenges and heightens the risk of truncated episodes, yielding sparser reward signals in training. Motivated by this rationale, we assess GARDEN across a set of generated graphs with diverse sparsity levels. We randomly generate $10$ graph topologies for $\beta \in \{0.01, 0.05, 0.1, 0.2, 0.3, 0.4, 0.5, 0.75, 1.0\}$ with number of nodes $n=200$ and $\alpha = 30$. We train GARDEN separately for each value of $\beta$ and gauge its performance against the baselines.

\subsection{Ablation of Node Representation}
We assess our GNN-based model against alternative designs through an ablation study on synthetic graphs. Using five random seeds and fixed graph parameters ($n=200$, $\alpha=30$, $\beta=5$), we conduct experiments on our three model designs: the MLPA2C model using only the raw node attributes $\mathbf{x}$, the MLPA2C variant incorporating both node attributes and degrees $\mathbf{x}^\text{WD}$, and the proposed GNN-based GARDEN method, which employs learned graph embeddings $\mathbf{x}^\text{GAT}$.

\section{Experimental Results}
\subsection{Facebook Graphs}

\begin{table}[t]
\caption{Metrics obtained by the methods on the 5 social network ego graphs. GARDEN consistently yields the best performance, followed by the DistanceWalker method.}
\label{table:fbgraphs}
\begin{center}
\resizebox{0.8\linewidth}{!}{
\begin{tabular}{lccccc}
\toprule
Metric & GreedyWalker & DistanceWalker & ConnectionWalker & RandomWalker & GARDEN (Ours) \\
\toprule
$\bar{R}_{\text{oracle}} (\downarrow)$ & 27.17 ± 1.09 & 15.98 ± 0.79 & 33.39 ± 1.15 & 33.17 ± 1.16 & \bf{10.95} ± 0.76 \\
 & 31.78 ± 1.22 & 13.01 ± 0.90 & 34.62 ± 1.17 & 34.27 ± 1.18 & \bf{9.89} ± 0.78 \\
 & 27.35 ± 1.19 & \bf{9.99 ± 0.78} & 38.84 ± 1.42 & 38.88 ± 1.43 & \bf{8.74 ± 0.76} \\
 & 25.69 ± 0.88 & 14.33 ± 0.65 & 28.25 ± 1.00 & 28.01 ± 1.01 & \bf{12.08 ± 0.71} \\
 & 28.29 ± 0.66 & 22.71 ± 0.70 & 28.11 ± 0.69 & 28.22 ± 0.67 & \bf{16.97 ± 0.74} \\
\midrule
$R_{\text{trunc}}(\downarrow)$ & 79.00 & 39.70 & 76.90 & 76.70 & \bf{14.90} \\
& 78.00 & 19.50 & 68.50 & 64.10 & \bf{12.60} \\
 & 70.00 & 13.90 & 69.80 & 67.20 & \bf{13.50} \\
 & 88.00 & 44.00 & 81.50 & 79.40 & \bf{40.10} \\
 & 96.00 & 72.70 & 89.50 & 90.00 & \bf{48.20} \\
\midrule
$R_{\text{win}}(\uparrow)$ & 7.90 & 26.20 & 2.70 & 2.50 & \bf{60.70} \\
 & 10.90 & 29.90 & 4.30 & 3.90 & \bf{51.00} \\
 & 11.90 & 33.80 & 1.50 & 2.00 & \bf{50.80} \\
& 6.10 & 32.10 & 5.30 & 8.20 & \bf{48.30} \\
 & 2.20 & 24.90 & 7.60 & 10.50 & \bf{54.80} \\ 
\end{tabular}
}
\end{center}
\end{table}

As shown in Table~\ref{table:fbgraphs}, we find that GARDEN significantly outperforms baselines across all the real-world datasets and metrics we have tested on. Given the variety of attribute dimensions and densities, as displayed in Table~\ref{tab:FBMetrics} in the Appendix, we may argue that in graphs with high amounts of latent structure, our model is robust to these factors. 

In Figure~\ref{fig:fbvis} in the Appendix, we visualize the value function learned by GARDEN on these social network ego graphs. This highlights that the values obtained by GARDEN serve as a reliable proxy for graph distance, assigning highest values to nodes in the target's cluster or clusters with strong connectivity to the target's community. Furthermore, it demonstrates the interpretability of the proposed technique for graph path search.

\chl{
In Table~\ref{tab:runtime} in the Appendix, we also include a runtime analysis to quantify the scalability of the proposed method. These results show that GARDEN maintains sub-millisecond per-action inference times even with CPU-only execution as the graph size increases, thanks to its decentralized nature. We therefore expect the method to maintain fast inference times even in substantially larger topologies.
}

\subsection{Sensitivity Analysis of Graph Density and Temperature Parameters}

\begin{wrapfigure}{R}{0.6\textwidth}
\vspace{-1\baselineskip}
    \centering
    \includegraphics[width=\linewidth]{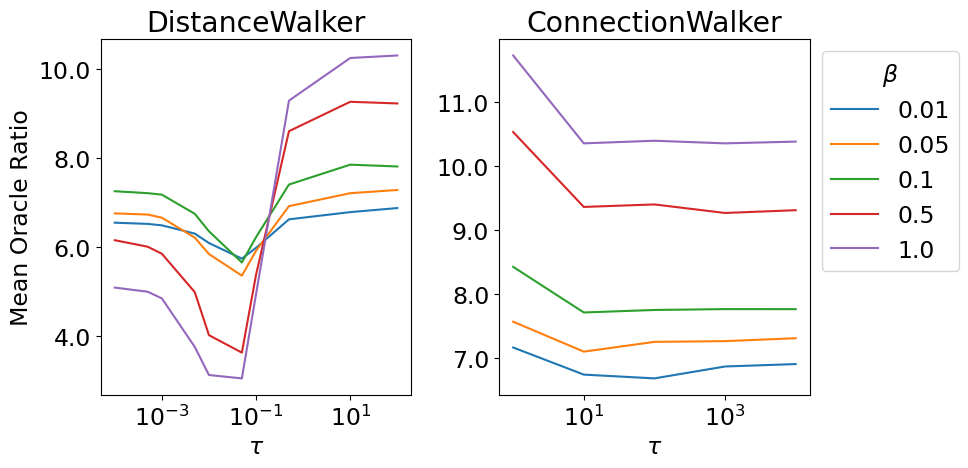}
    \caption{Mean Oracle Ratio obtained by the stochastic baselines on the validation dataset as a function of the temperature $\tau$ for varying values of $\beta$.}
    \label{fig:crossval}
\vspace{-1.5\baselineskip}
\end{wrapfigure}

In Figure~\ref{fig:crossval}, we show the validation performance as a function of the SoftMax temperature $\tau$ of the stochastic DistanceWalker and ConnectionWalker baselines. For both methods, a middle-ground temperature value yields shorter path lengths. Furthermore, performance is more sensitive to $\tau$ for the DistanceWalker method.

\begin{figure}[t]
    \centering
    \includegraphics[width=0.8\linewidth]{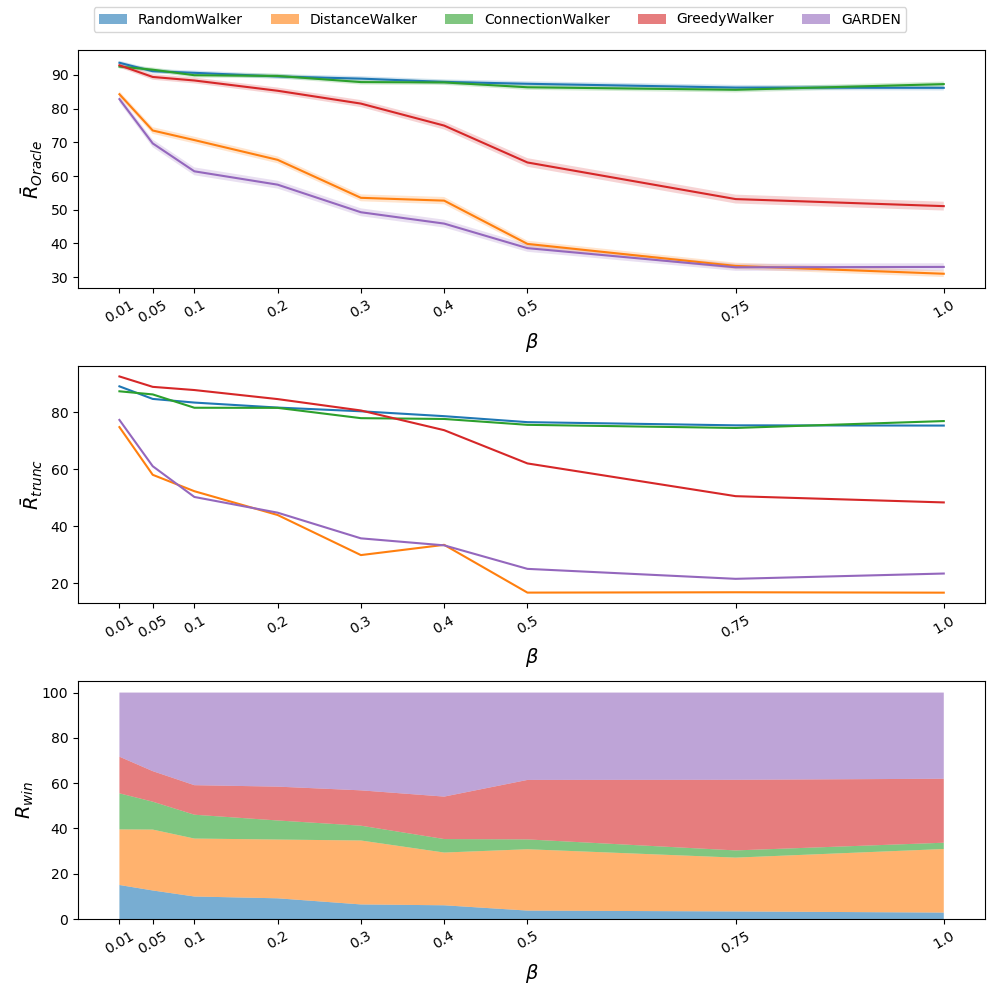}
    \caption{Metric values obtained by the methods as a function of the synthetic graph density parameter $\beta$. GARDEN generally performs best, but it is notably surpassed by DistanceWalker in the truncation rate for high values of $\beta$.}
    \label{fig:sensitivityanalysis}
\end{figure}

The sensitivity analysis for the synthetic graph density parameter $\beta$ is shown in Figure~\ref{fig:sensitivityanalysis}. GARDEN consistently matches or surpasses baseline performance for all $\beta$ values  for both Mean Oracle Ratio and Win Rate metrics. However, DistanceWalker outperforms our model for higher $\beta$ in Truncation Rate. In this setting, DistanceWalker benefits from knowledge of the ``true'' node attributes determining link generation and high $\beta$ values leading to most connections being realized. This is in contrast with the gap on real-world datasets, for which this metric is not available: indeed, GARDEN may be seen as recovering an underlying ``hidden metric''.

\subsection{Ablation of Node Representation}

\begin{wraptable}{R}{0.45\textwidth}
\vspace{-1.5\baselineskip}
\caption{Ablation results obtained by pairing different node representations with the proposed problem formulation and Reinforcement Learning algorithm.}
\label{table:ablation}
\begin{center}
\resizebox{\linewidth}{!}{

\begin{tabular}{llll}
\toprule
Agent & $\bar{R}_{\text{oracle}} (\downarrow)$ & $R_{\text{trunc}}(\downarrow)$ & $R_{\text{win}}(\uparrow)$ \\
\midrule
GARDEN & \bf{1.95±0.09} & \bf{1.68} & 34.44 \\
MLPA2CWD & 2.23±0.13 & 2.94 & 30.78 \\
MLPA2C & 2.31±0.12 & 4.32 & \bf{34.78} \\
\bottomrule
\end{tabular}
}
\end{center}
\end{wraptable}

As shown in Table \ref{table:ablation}, GARDEN significantly outperforms the MLPA2C and MLPA2CWD designs that only use raw node attributes and MLP-parameterized policies in both $R_{\text{trunc}}$ and $\bar{R}_{\text{oracle}}$. The standard MLP-parameterized model achieves the best Win Rate $R_{\text{win}}$, and the differences with respect to this metric are less conclusive. This can be explained by the arbitrary tie-breaking performed when path lengths match for all methods, coupled with the high density parameter $\beta=5$ leading to shorter path lengths.

\section{Conclusions and Future Work}

In this paper, we have considered the problem of decentralized search in graphs, which is motivated by privacy, scalability, and dynamicity requirements of many network modeling scenarios. Despite the lack of a central view of the network, the homophily and community structure observed in many networks can allow for decentralized agents to find short paths to a given target, as famously demonstrated by the Milgram experiment~\cite{travers1969}.

We have proposed the GARDEN method to address this problem, which trains agents in a centralized fashion and allows for decentralized policy execution. Our approach is based on message routing policies that are learned using Reinforcement Learning, paired with node features learned by a Graph Neural Network specifically for the task. Our results show that our method can outperform stochastic routing policies based on attribute or structural information alone. 
It is possible to observe significant improvements when searching on real-world social network graphs with non-trivial latent structures and high-dimensional node attributes.

For simplicity, we have considered a \textit{memory-less} search procedure that is akin to a biased random walk. This means that the agents cannot react to the unsuccessful exploration of a given region of the graph before arriving at a previously visited node, and the same distribution over actions will apply independently of the historical trajectory. The problem formulation can be extended by including the history of visited nodes in the message $\mathbf{m}$ and forbidding using already-visited nodes as actions. RNNs may be used to encode the history as input to the learned models, as performed in other learning-based graph search works~\cite{shen2018m,pandy2022learning}.

We believe that our results provide evidence for a sort of ``hidden metric'' hypothesis, showing how a latent feature space amenable for graph navigation can be recovered by reward-driven learning. An interesting aspect that can be considered by future work is to compare these emergent representations with the means in which individuals take decisions for routing messages in experiments conducted over real social networks.

\bibliographystyle{unsrtnat}
\bibliography{reference}

\appendix
\section{Appendix}
\setcounter{topnumber}{3}

\subsection{Implementation Details}

\chl{Our implementation is publicly available at \url{https://github.com/flxclxc/rl-graph-search}}. Please see the \texttt{README.md} file for instructions on how to set up the dependencies, download the publicly available data, and run the code. 

We train our models using the Adam optimizer \cite{kingma2014adam} for $200,000$ episodes, evaluating performance every $100$ episodes on the serialized validation set. Early stopping is applied based on the validation loss $\bar{R}_{\text{oracle}}$. Unless otherwise stated, we train and evaluate models over $10$ random seeds, reporting confidence intervals where appropriate. Table \ref{table:config} presents the hyperparameter configuration shared across the three model designs. We fix $\gamma=0.99$ and the maximum episode length $T_\text{max}=100$. Lastly, when providing node input features to the GAT, we append a binary indicator variable that signals that a particular node $u$ is the center of the ego graph to the raw attributes defined as $[\mathbf{x}_{w} || \mathbb{I}[w=u]]$. 
This is used to distinguish the node from which the message must be sent.

\chl{We note that, while the GAT-based model contains 3 layers, we only use the ego graphs centered around the nodes to compute the embeddings, hence ensuring that only 1-hop visibility is provided. The first layer is fed the raw node attributes, while the second and third layers use the ``latent'' embeddings constructed by the previous layer. The message-passing therefore occurs over the same 1-hop subgraph in all the layers.
}

\subsection{Additional Tables and Figures}

\chl{
\textbf{Summary statistics.} Statistics about the considered real-world graphs are shown in Table~\ref{tab:FBMetrics}.

}

\chl{
\textbf{Runtime analysis}. We carry out a runtime analysis to examine the scalability and computational cost of GARDEN. In terms of methodology, we calculate the mean action time, i.e., the elapsed wall clock time measured in milliseconds from the arrival of a message at a node until an action is chosen. The measurements are averaged over $100$ target nodes and are carried out using an Intel i7-11800H CPU. We note that the execution of GARDEN also involves a time overhead for creating the local ego graph embeddings from $f_{\text{rep}}^{\Theta_3}$, which is reported separately.}

\begin{table}[h]
    \centering
    \resizebox{0.5\linewidth}{!}{
        \begin{tabular}{lc}
            \toprule
            Parameter & Value \\ 
            \midrule
            Actor Network Hidden Dimensions & 64 \\ 
            Actor Network Layers & 3 \\ 
            Critic Network Hidden Dimensions & 64 \\ 
            Critic Network Layers & 3 \\ 
            Graph Attention Network Hidden Dimensions & 64 \\ 
            Graph Attention Network Heads & 1 \\
            Graph Attention Network Layers & 3 \\ 
            Entropy Regularization Coefficient $\lambda$ & $1 \times 10^{-3}$ \\ 
            \bottomrule
        \end{tabular}
    }
    \caption{Hyperparameter configuration used for all the learning-based models.}
    \label{table:config}
\end{table}

\begin{table}[t]
    \centering 
    \resizebox{0.3\linewidth}{!}{
    \begin{tabular}{@{}cccccc@{}}
    \toprule 
    Graph & $n$ & $l_{G}$ & $\rho$ & $d$ \\
    \midrule
    1 & 148 & 2.69 & 0.16 & 105 \\
    2 & 168 & 2.43 & 0.12 & 63 \\
    3 & 224 & 2.52 & 0.13 & 161 \\
    4 & 324 & 3.75 & 0.05 & 224 \\
    5 & 532 & 3.45 & 0.03 & 262 \\
    \bottomrule
    \end{tabular}
    }
    \caption{Number of nodes $n$, average shortest path length $l_{G}$, edge density $\rho$ and attribute dimension $d$ for each real-world ego graph.}
\label{tab:FBMetrics}
\end{table}

\begin{table}[t]
\centering 
\chl{\resizebox{0.8\linewidth}{!}{

\begin{tabular}{lrrrrr}
\toprule
Graph & 1 & 2 & 3 & 4 & 5 \\
\midrule
Number of Nodes & 148 & 168 & 224 & 324 & 532 \\
Avg \# Neighbors & 22.86 & 19.71 & 28.5 & 15.52 & 18.09 \\
Overhead Time GARDEN (ms) & 0.3024 & 0.2349 & 0.5024 & 0.2214 & 0.2143 \\
Action Time: GARDEN (ms) & 0.1592 & 0.1715 & 0.1800 & 0.1602 & 0.1656 \\
Action Time: MLPA2CWD (ms) & 0.1446 & 0.1548 & 0.1686 & 0.1532 & 0.1520 \\
Action Time: MLPA2C (ms) & 0.1499 & 0.1487 & 0.1518 & 0.1464 & 0.1469 \\
Action Time: GreedyWalker (ms) & 0.0259 & 0.0256 & 0.0323 & 0.0281 & 0.0305 \\
Action Time: DistanceWalker (ms) & 0.0657 & 0.0697 & 0.0735 & 0.0674 & 0.0685 \\
Action Time: ConnectionWalker (ms)& 0.0453 & 0.0522 & 0.0562 & 0.0456 & 0.0517 \\
Action Time: RandomWalker (ms)& 0.0006 & 0.0006 & 0.0007 & 0.0006 & 0.0006 \\
\bottomrule
\end{tabular}
}
}
\caption{\chl{Runtime analysis comparing average time for calculating actions across all models, including overhead computation time of local graph embeddings for GARDEN.}}
\label{tab:runtime}
\end{table}

\chl{The results are shown in Table~\ref{tab:runtime}. They demonstrate that both the embedding overhead time and action time increase slightly as the number of neighbors grows, but stays reasonably bounded. As it is expected, a clear hierarchy is present in which the neural network-based models require more inference time, followed by the heuristic attribute-based baselines, and finally the simple random walk baseline. This analysis highlights the fact that GARDEN maintains sub-millisecond inference times even with CPU-only execution as the graph size increases, thanks to its decentralized nature. Notably, the total inference time is lower on the largest 532-node graph compared to the smallest 148-node graph, owing to differences in density. We therefore expect the method to maintain fast inference times even in substantially larger topologies.
}

\begin{figure}[th]
    \centering
    \includegraphics[width=\linewidth]{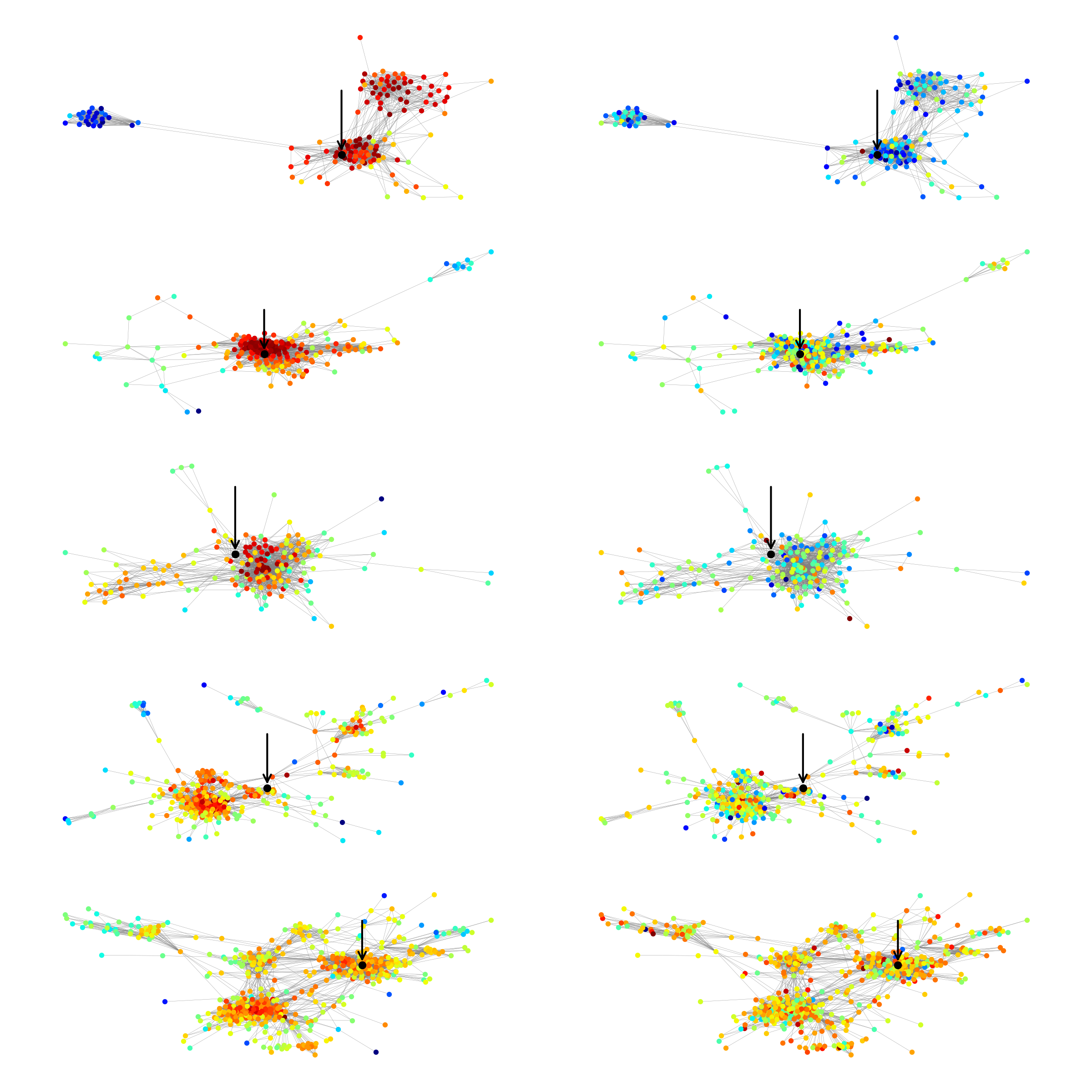}
\caption{Visualization of the learned value function $v(u, u_\text{tgt})$ learned by GARDEN for each node $u$ (left) and preferability score $-\|\mathbf{x}_{u} - \mathbf{x}_{u_\text{tgt}}\|_{2} / \tau$ of the DistanceWalker baseline (right) for the social network graphs. The black arrows indicate the target node, while the color intensities of the other nodes are proportional to the value function learned by GARDEN (left) or baseline score (right). Concretely, dark red nodes indicate high proximity to the target, while dark blue nodes reflect low proximity. GARDEN recovers meaningful values for performing graph navigation, effectively leveraging proximity in both node attributes and topological structure.}
    \label{fig:fbvis}
\end{figure}

\chl{
\textbf{Interpretability of the learned value function.}} In Figure~\ref{fig:fbvis}, we plot GARDEN's learned value function \chl{$f_{v}^{\Theta_2}$} across the social network ego graphs. Brighter colors indicate a higher estimated value function relative to the target node, which is indicated with a black arrow and chosen randomly from the respective test sets. For comparison, we also plot the implicit preferability score $-\|\mathbf{x}_{u} - \mathbf{x}_{u_{\text{tgt}}}\|_{2} / \tau$ generated by the best-performing baseline, DistanceWalker, for the same source-target pairs. DistanceWalker struggles with Euclidean pairwise attribute distance due to high dimensionality and sparsity of node attributes. Conversely, values obtained by GARDEN serve as a reliable proxy for graph distance, assigning highest values to nodes in the target's cluster or clusters with strong connectivity to the target's community.

\end{document}